\documentclass[10pt,twocolumn,letterpaper]{article}

\usepackage{cvpr}              

\usepackage{makecell}
\usepackage{dsfont}

\usepackage{amssymb}
\usepackage{pifont}
\definecolor{cvprblue}{rgb}{0.21,0.49,0.74}
\usepackage[pagebackref,breaklinks,colorlinks,allcolors=cvprblue]{hyperref}

\def\paperID{*****} 
\def\confName{CVPR}
\def\confYear{2026}

\title{PDF-GS: Progressive Distractor Filtering for Robust 3D Gaussian Splatting
}

\author{
Kangmin Seo,
MinKyu Lee,
Tae-Young Kim,
ByeongCheol Lee,
JoonSeoung An,
Jae-Pil Heo\footnotemark\\
Sungkyunkwan University\\
{\tt\small \{skmskku, bluelati98, jackdawson, bc7817, ajs3801, jaepilheo\}@skku.edu}
}

\usepackage{graphicx}
\usepackage{booktabs}
\usepackage{multirow}
\usepackage[table]{xcolor}
\usepackage{siunitx}
\usepackage{adjustbox}
\usepackage{caption}
\usepackage{placeins}

\definecolor{col3view}{RGB}{220,235,255}
\definecolor{col6view}{RGB}{230,255,230}
\definecolor{col9view}{RGB}{255,240,225}
\newcolumntype{B}{>{\columncolor{col3view}}c}
\newcolumntype{G}{>{\columncolor{col6view}}c}
\newcolumntype{P}{>{\columncolor{col9view}}c}

\definecolor{bestbg}{RGB}{243,185,184}
\definecolor{secbg}{RGB}{255,225,196}
\definecolor{thirdbg}{RGB}{255,251,206}
\newcommand{\best}[1]{\cellcolor{bestbg}#1}
\newcommand{\secondbest}[1]{\cellcolor{secbg}#1}
\newcommand{\thirdbest}[1]{\cellcolor{thirdbg}#1}

\let\bo\best
\let\bs\secondbest
\let\bt\thirdbest
\newcommand{\bn}[1]{#1}

\begin{document}

\twocolumn[{
\maketitle
\begin{center}
    \centering
    \captionsetup{type=figure}
    \includegraphics[width=\textwidth]{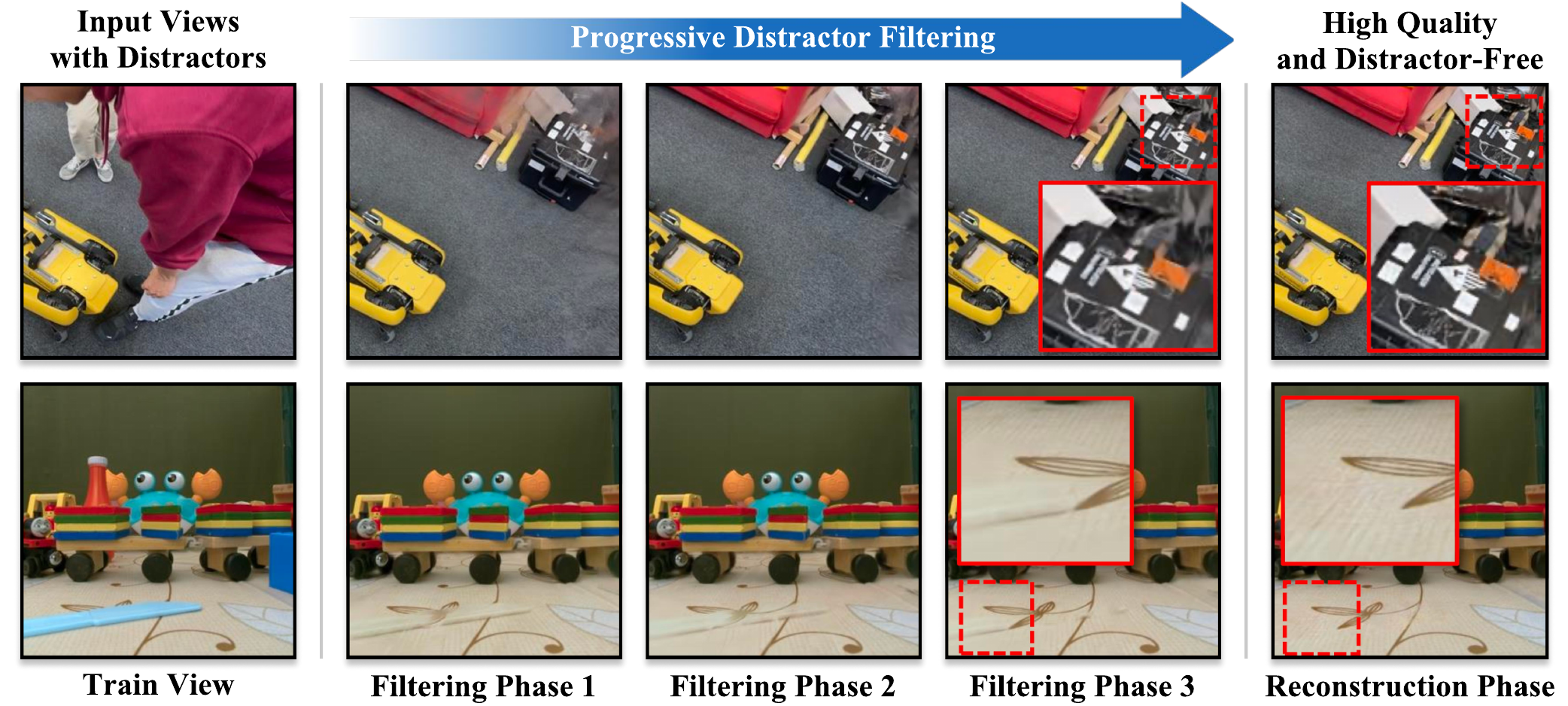}
\caption{\textbf{Overview of PDF-GS.} 
During the \textit{Progressive Filtering Phases} (phases 1–3), PDF-GS progressively removes transient, view-inconsistent distractors. 
Across successive phases, inconsistent regions are suppressed while stable, view-consistent structures are preserved. 
In the final \textit{Reconstruction Phase} (phase 4), fine-grained appearance details are recovered from the purified representation, leading to a high-fidelity and distractor-free 3D reconstruction.}
  \label{fig:motivation}
\end{center}%
}]

\renewcommand{\thefootnote}{}
\footnotetext{$^{*}$Corresponding author}
\renewcommand{\thefootnote}{\arabic{footnote}}

\begin{abstract}
Recent advances in 3D Gaussian Splatting (3DGS) have enabled impressive real-time photorealistic rendering. However, conventional training pipelines inherently assume full multi-view consistency among input images, which makes them sensitive to distractors that violate this assumption and cause visual artifacts. In this work, we revisit an underexplored aspect of 3DGS: its inherent ability to suppress inconsistent signals. Building on this insight, we propose PDF-GS (Progressive Distractor Filtering for Robust 3D Gaussian Splatting), a framework that amplifies this self-filtering property through a progressive multi-phase optimization.
The progressive filtering phases gradually remove distractors by exploiting discrepancy cues, while the following reconstruction phase restores fine-grained, view-consistent details from the purified Gaussian representation. Through this iterative refinement, PDF-GS achieves robust, high-fidelity, and distractor-free reconstructions, consistently outperforming baselines across diverse datasets and challenging real-world conditions. Moreover, our approach is lightweight and easily adaptable to existing 3DGS frameworks, requiring no architectural changes or additional inference overhead, leading to a new state-of-the-art performance. The code is publicly available at \url{https://github.com/kangrnin/PDF-GS}.
\end{abstract}

\section{Introduction}

Recent advances in 3D Gaussian Splatting (3DGS)~\cite{3dgs} have demonstrated impressive rendering quality and efficiency for novel view synthesis~\cite{gaussianslam, langsplat, test-to-3d-gs, gaussianedit}. However, standard 3DGS pipelines operate under an implicit assumption that all training images depict a static and view-consistent scene. In real-world environments, this assumption is often violated. Scenes frequently include \textit{distractors}, such as transient, dynamic, or view-dependent contents (e.g., pedestrians, vehicles, or shadows). These distractors introduce multi-view inconsistencies that corrupt the training signal, resulting in artifacts, blurred regions, and unstable geometry in the reconstructed 3D representation.

\begin{figure}[t]
  \centering
  \begin{subfigure}[t]{0.48\columnwidth}
    \centering
    \includegraphics[width=\linewidth, trim=60 0 0 100, clip]{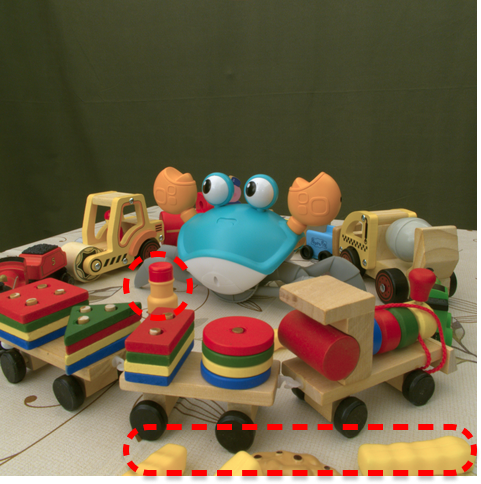}
    \caption{Ground Truth}
  \end{subfigure}\hfill
  \begin{subfigure}[t]{0.48\columnwidth}
    \centering
    \includegraphics[width=\linewidth, trim=60 0 0 100, clip]{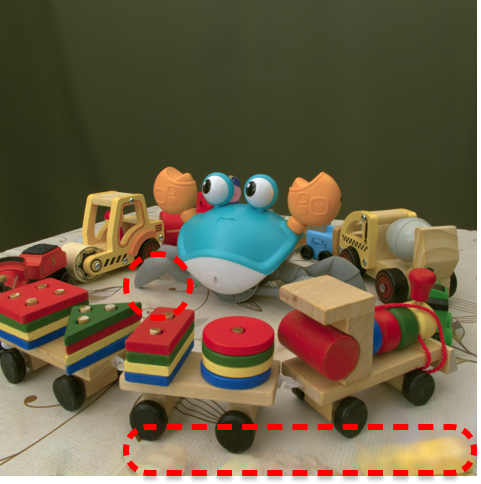}
    \caption{3DGS Render}
  \end{subfigure}\hfill
  \vspace{-5pt}
    \caption{\textbf{Self-filtering behavior of vanilla 3DGS.} 
    After standard 3DGS training, distractor objects visible in the train view (a) are removed in the rendered result (b), indicating that inconsistent regions tend to either disappear or become blurred in the reconstruction (\textit{Crab2} scene in RobustNeRF dataset~\cite{robustnerf}).
    }
  \label{fig:figure1}
\end{figure}

Prior approaches have primarily relied on explicit distractor modeling through mask prediction or decomposition~\cite{spotless,robustsplat,desplat,hybridgs}.
In this work, we take a conceptually distinct perspective by leveraging a fundamental yet underexplored property of 3DGS: its inherent ability to suppress view-inconsistent regions while faithfully reconstructing view-consistent structures. Transient or view-dependent objects, often regarded as distractors, fail to provide consistent multi-view observations and therefore tend to disappear or become blurred in rendered outputs (Fig.~\ref{fig:figure1}). We refer to this observation as the self-filtering phenomenon, which motivates our key idea: to reinterpret 3DGS as a refinement mechanism that removes distractors from the training data.

However, this self-filtering behavior alone is not sufficient to eliminate distractors completely. In practice, distractor-driven artifacts often remain after a single optimization pass (Fig.~\ref{fig:motivation}). Motivated by this limitation, we aim to progressively amplify 3DGS’s intrinsic filtering behavior through iterative refinement. To this end, we propose PDF-GS, a multi-phase framework comprising (1) Progressive Filtering Phases and a subsequent (2) Reconstruction Phase.

\textbf{Progressive Filtering Phases} build on the self-filtering behavior of 3DGS, reinterpreting it as a mechanism for distractor removal. Their primary objective is to iteratively refine the scene by identifying and eliminating distractors, rather than fine-grained reconstruction. Each filtering phase aims to refine the scene representation by gradually reducing the influence of distractors while preserving view-consistent structures. Specifically, each filtering phase first localizes and masks out distractor-prone regions in the train image, based on the discrepancy between the train image and the rendered outputs from the previous phase. Each filtering phase then introduces a re-initialized set of Gaussian parameters, and optimizes it with the masked train images, where the newly optimized parameters are passed to the next phase. By iteratively alternating between discrepancy-based masking and re-optimization, the process progressively exposes and suppresses distractors that cannot be reliably reconstructed, producing increasingly purified and distractor-free Gaussian parameter sets.

The \textbf{Reconstruction Phase} aims to recover fine-grained details that are insufficiently optimized during the preceding filtering phases. In this stage, we apply the standard 3DGS loss to refine the geometry and appearance of the scene for fine-grained details, while maintaining the discrepancy-based distractor mask to prevent the reactivation of inconsistent regions. The result from this phase serves as the final 3DGS model of our entire framework.

Through this multi-phase design, PDF-GS reframes 3DGS as an active \textit{inconsistency filter} and subsequently leverages it for high-fidelity reconstruction. Our approach is conceptually simple, requires no architectural modifications to 3DGS, does not require any computational overhead at inference time.

\section{Related Works}
\subsection{Novel View Synthesis}
Novel View Synthesis is the task of producing photorealistic views from novel viewpoints given a limited set of posed images. 
Neural Radiance Fields (NeRF)~\cite{nerf}, which represent a scene as a coordinate-based volumetric radiance field and enable MLP-based differentiable rendering via ray marching, has led to a wide range of methods for novel-view synthesis~\cite{deblurnerf,mipnerf,D-nerf, NERFwild}. However, because NeRF employs an MLP-based implicit representation, it suffers from slow rendering speed. To overcome this limitation, 3D Gaussian Splatting (3DGS)~\cite{3dgs} has recently been proposed, which represents a scene as an explicit set of anisotropic 3D Gaussians and renders them via differentiable splatting, enabling real-time rendering~\cite{mipsplatting, vastsplat, sparsegs, gaussianedit}.

\begin{figure*}[t]
    \centering
    \includegraphics[width=\textwidth]{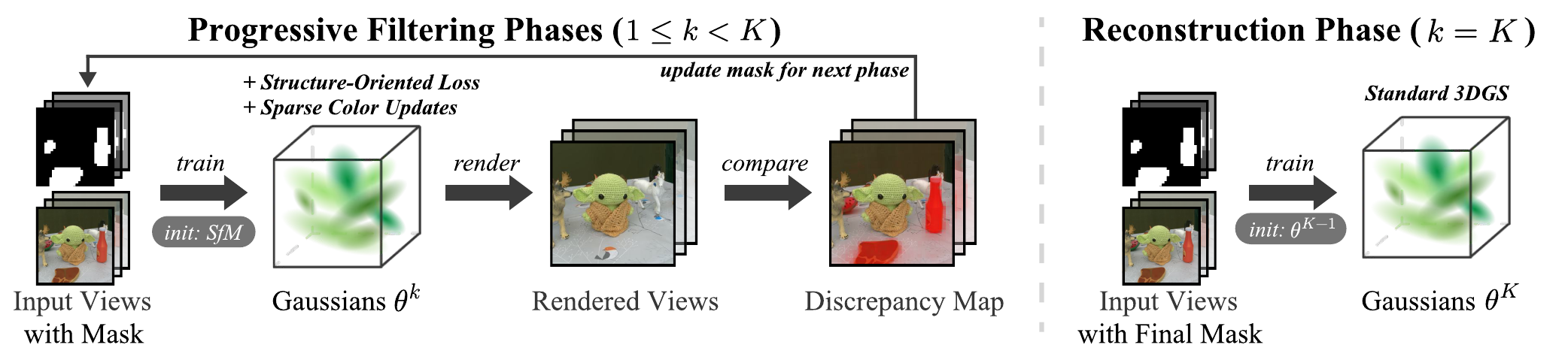}
    \caption{\textbf{Conceptual illustration of PDF-GS.} 
    Our method progressively filters out transient and view-inconsistent distractors through iterative refinement. 
    During the filtering phases, regions exhibiting multi-view inconsistencies are identified by the discrepancies between rendered views and training images. 
    These inconsistent regions are then masked out, while stable view-consistent regions exhibiting small discrepancy are preserved. This progressive filtering process yields a purified 3D Gaussian representation, but often lacks fine-grained details. In the final reconstruction phase, we aim to further restore the fine-grained details that are not fully optimized during the filtering phases. Accordingly, we re-introduce the standard 3DGS loss which leads to both high-fidelity and distractor-free scene reconstruction. The output of the reconstruction phase serves as the final result of our framework.}
    \label{fig:method}
\end{figure*}

\subsection{Robustness Against Distractors in 3DGS}
While methods for novel view synthesis~\cite{nerf,3dgs} assume a static scene, this assumption fails in unconstrained real-world captures where dynamic or transient objects (distractors) are often included, leading to inaccurate scene reconstructions. Recently, many 3DGS-based approaches have been proposed to address dynamic/transient objects. SpotLessSplats~\cite{spotless} leverages pretrained semantic features from Stable Diffusion~\cite{stablediffusion} and detects distractors using spatial and spatio-temporal clustering combined with a robust masking scheme. DeSplat~\cite{desplat} decomposes a 3DGS scene into static Gaussian space and per-view transient Gaussian space by optimizing only a photometric loss, without relying on a pretrained model. HybridGS~\cite{hybridgs} shares motivation with DeSplat, but detects distractors using 2D Gaussians~\cite{gaussianimage} and a multi-stage training scheme. RobustSplat~\cite{robustsplat} mitigates artifacts caused by distractors in 3DGS by delaying densification and applying a coarse-to-fine bootstrapping of transient masks using DINOv2~\cite{dinov2} features. AsymGS~\cite{asymgs} trains two 3DGS models in parallel, leveraging cross-model consistency to suppress stochastic artifacts.
In this work, we take a different perspective by leveraging the inherent property of 3DGS to preserve multi-view consistency during optimization, where we progressively identify and remove distractors and achieve robust and high-fidelity reconstruction.

\section{Method}

\noindent\textbf{Overview.} The key motivation of PDF-GS is the observation of the inherent property of 3DGS: its tendency to suppress inconsistent signals and naturally filter out distractors (Fig.~\ref{fig:figure1}). However, we also find that a single optimization pass often fails to fully suppress distractor-driven artifacts. Building on this observation, we introduce a multi-phase framework consisting of:
\textbf{(1)} \textit{Progressive Filtering Phases} and
\textbf{(2)} a consecutive \textit{Reconstruction Phase}.

The \textit{Progressive Filtering Phases} operate over refinement phases indexed by $k$, where $1 \le k < K$. The primary goal of these phases is to identify and eliminate distractor signals, focusing on purification instead of detailed reconstruction. Each refinement phase produces a Gaussian parameter set $\Theta^{(k)}$ that becomes progressively purified of distractors. After $K-1$ phases, the process transitions to the Reconstruction Phase at phase $k = K$.

The \textit{Reconstruction Phase} then generates the final 3DGS representation, recovering fine-grained details that may be attenuated during the progressive filtering phases. Together, these phases yield a 3D representation that maintains stable multi-view structural consistency while producing distractor-free, high-fidelity reconstructions.

\subsection{Progressive Filtering Phases}

\noindent\textbf{Overview.} We begin with a converged initial Gaussian parameter set $\Theta^{(1)}$ for $k=1$.
Phase $k$ proceeds through two key steps:
(1) identifying distractor regions using the previous representation $\Theta^{(k-1)}$, and
(2) training a new parameter set $\Theta^{(k)}$, which is re-initialized from Structure-from-Motion (SfM) points, while masking out those identified distractor regions from the training images.
By repeatedly updating the discrepancy maps and re-optimizing with masked supervision, each phase progressively improves the fidelity and robustness of the Gaussian representation.

\begin{figure}[t]
    \vspace{-5pt}

    \centering
    \captionsetup[sub]{font=scriptsize}
    \begin{subfigure}[t]{0.32\columnwidth}
    \centering
    \includegraphics[width=\linewidth, trim={80 0 80 0}, clip]{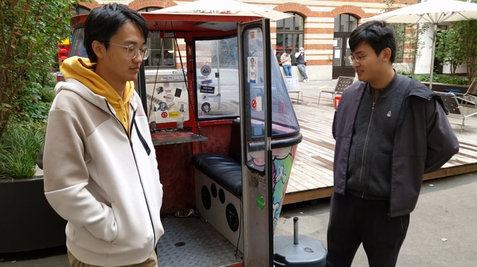}
    \caption{Ground Truth}
    \end{subfigure}\hfill
    \centering
    \begin{subfigure}[t]{0.32\columnwidth}
    \centering
    \includegraphics[width=\linewidth, trim={80 0 80 0}, clip]{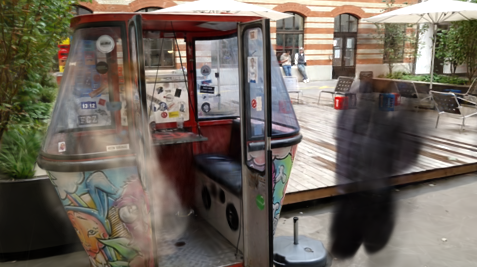}
    \caption{Standard 3DGS Loss}
    \end{subfigure}\hfill
    \begin{subfigure}[t]{0.32\columnwidth}
    \centering
    \includegraphics[width=\linewidth, trim={80 0 80 0}, clip]{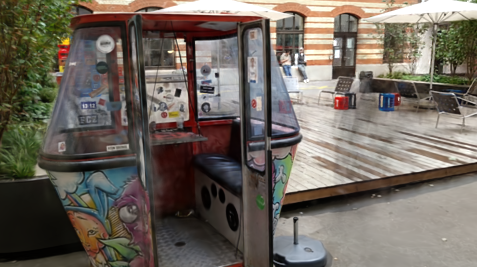}
    \caption{Structure-Oriented}
    \end{subfigure}
    \vspace{-5pt}
    \caption{
    \textbf{Effectiveness of the structure-oriented objective (Eq.~\ref{ssim}).
    }
    The model trained with standard 3DGS loss (b) overfits transient distractors and exhibits color bleeding, while using structure-oriented objective (c) preserves structural consistency and better suppresses distractors (\textit{Patio} scene in NeRF~On-the-go~\cite{nerf-on-the-go}).}
    \label{fig:ssim_ablation}

\end{figure}

\subsubsection{Initialization \texorpdfstring{$\boldsymbol{(k=1)}$}{(k=1)}.}
We begin the refinement process by constructing an initial set of Gaussian parameters $\Theta^{(1)}$ that serves as a coarse but reliable starting point.
This initialization leverages the natural self-filtering behavior of 3DGS, which tends to suppress multi-view inconsistent content~(Fig.~\ref{fig:figure1}).
Accordingly, obtaining $\Theta^{(1)}$ is straightforward: we largely follow the standard 3DGS training procedure but with modifications specified in Sec.~\ref{sec:filtering_train_objective} for additional robustness. The resulting Gaussian representation is expected to capture the dominant, view-consistent structure of the scene.

\subsubsection{Progressive Filtering \texorpdfstring{$\boldsymbol{(1 < k < K)}$}{(1 < k < K)}.}
\noindent\textbf{Remark.} Within the Progressive Filtering stages, we aim to identify and mask out distractor regions in the training images, thereby progressively improving the robustness of the Gaussian parameter $\Theta^{(k)}$ for each phase $k$. Note that at each progressive filtering phase, we re-initialize a fresh Gaussian set from the SfM points, instead of reusing the Gaussians from the prior phase. This design prevents accumulated error propagation of prior steps such as color drift or geometric biases (Fig.~\ref{fig:filtering}).

\vspace{5pt}
\noindent\textbf{Distractor Identification.}
At each filtering phase $k$, distractor regions are identified by comparing the training images with the rendered outputs generated from the previous Gaussian parameter set $\Theta^{(k-1)}$.
The intuition is that 3DGS naturally suppresses signals that are not consistent across views. Transient or view-dependent content, i.e., distractors, is not stably reconstructed and thus tends to diminish in the rendered outputs. As a result, these regions exhibit significant discrepancies when compared against the corresponding training images, enabling reliable identification of distractor areas.

Specifically, for training view $v \in \mathcal{V}$ and refinement phase $k$, we compute a discrepancy map that measures the difference between the ground truth image $I^{gt}_v$ and the rendered image $I^{r}_v(\Theta^{(k-1)})$ from the Gaussians of the previous phase $k-1$. With an additional transformation $F(\cdot)$ for robustness under low-level visual variation, the discrepancy map is defined as:
\begin{equation}
D^{(k-1)}_v
= \left\| F(I^{gt}_v) - F\left(I^{r}_v(\Theta^{(k-1)})\right) \right\|_2.
\label{Eq: discrepancy metric}
\end{equation}
Here, we use DINOv3~\cite{dinov3} features for $F(\cdot)$, yielding a patch-level semantic discrepancy measure.
Because $\Theta^{(k-1)}$ becomes increasingly purified across phases, the resulting $D^{(k-1)}_v$ provides progressively more accurate localization of distractors, forming the basis for the subsequent masked optimization step.

\vspace{5pt}
\noindent\textbf{Progressive Masked Optimization.}
Regions with high discrepancy are interpreted as distractors and thereby should be removed from the training signal, while the remaining regions provide clean structural supervision. Accordingly, given the discrepancy map $D^{(k-1)}_v$, we derive a binary mask that excludes distractor regions from supervision during the current filtering phase.  
For each pixel $p$ in view $v$, the mask and the corresponding masked images are defined as:
\begin{equation}
    M^{(k-1)}_v(p) = \mathds{1}\!\left[\, D^{(k-1)}_v(p) \le \tau_k \,\right],
    \label{maskgen}
\end{equation}
\begin{equation}
    \tilde{I}^{gt,(k)}_v = M^{(k-1)}_v \odot I^{gt}_v,\;
    \tilde{I}^{r,(k)}_v = M^{(k-1)}_v \odot I^{r}_v(\Theta^{(k)}),
\end{equation}
where $\tau_k$ is the threshold at phase $k$ and $\mathds{1}$ is the indicator function. Using the masked images $\tilde{I}^{gt,(k)}_v$ and $\tilde{I}^{r,(k)}_v$, the current Gaussian parameter set $\Theta^{(k)}$ is optimized following the standard 3DGS training pipeline, but with robustness-oriented modifications as specified in Sec.~\ref{sec:filtering_train_objective}.  

Importantly, this masked optimization progressively improves over phases: as $\Theta^{(k-1)}$ becomes more refined, discrepancy maps $D^{(k-1)}_v$ and the corresponding masks $M^{(k-1)}_v$ can better identify true distractor regions.  Consequently, each filtering phase $k$ benefits from a more distractor-free supervision signal than the previous one, enabling the Gaussian representation $\Theta^{(k)}$ to become more robust and structurally consistent.

\subsubsection{Training Objective \texorpdfstring{$\boldsymbol{(1 \le k < K)}$}{(1 ≤ k < K)}}
\label{sec:filtering_train_objective}

\noindent\textbf{Remark.} Below we elaborate on two modifications introduced on top of the standard 3DGS framework.
These modifications are designed not to enhance fine-grained appearance modeling, but rather to reshape the optimization trajectory such that distractor-prone, multi-view inconsistent regions become clearly excluded during training. As a result, subtle low-level appearance differences may arise compared to standard 3DGS optimization; however, when coupled with an appropriate metric robust under subtle low-level variations (e.g., Eq.~\ref{Eq: discrepancy metric} with DINOv3 features), these modifications significantly improve robustness in identifying and suppressing distractors. These modifications are employed during the filtering phases (for $1 \leq k < K$).

\vspace{5pt}
\noindent\textbf{Structure-Oriented Supervision.}
To achieve robustness under challenging, distractor-heavy training conditions, we adopt a purely structural supervision signal rather than conventional 3DGS loss. This choice is motivated by observations in prior work~\cite{nerf-on-the-go}, which indicate that structural similarity measures such as SSIM~\cite{SSIM} emphasize patch-level geometric consistency while being comparatively insensitive to localized color variations or transient artifacts.  
This property aligns well with our goal of suppressing distractors, which are typically view-inconsistent and often manifest through unstable color or texture patterns.

For filtering phase $k$, the training objective within the filtering phases $L^{(k)}_{\text{filter}}$ is defined as:

\begin{equation}
\mathcal{L}^{(k)}_{\text{filter}} =
\begin{cases}
\sum_{v \in \mathcal{V}}
\mathcal{L}_\text{ssim}\left({I}^{r,(k)}_v, {I}^{gt,(k)}_v\right), & \!\!\! k = 1, \\
\sum_{v \in \mathcal{V}}\mathcal{L}_\text{3DGS}\left(\tilde{I}^{r,(k)}_v, \tilde{I}^{gt,(k)}_v\right), & \!\!\! \text{otherwise,}
\end{cases}
\label{ssim}
\end{equation}
where $\mathcal{L}_\text{3DGS}$  refers to the standard 3DGS loss~\cite{3dgs} which consists of an L1 loss and an SSIM loss.

In the first filtering phase, by relying solely on structural similarity, this structure-oriented loss encourages the model to preserve globally consistent geometry while avoiding premature overfitting to view-dependent distractors.  
Empirically, we find that structural-only supervision plays a critical role in stabilizing the filtering process and preventing distractor accumulation during early stages of optimization (Fig.~\ref{fig:ssim_ablation}).

\vspace{5pt}

\vspace{5pt}
\noindent\textbf{Sparse Color Updates for View Consistency.}
SfM-initialized Gaussians provide a reliable starting point with stable geometry and view-consistent colors, as SfM inherently relies on cross-view feature correspondences. However, during the filtering process, frequent color updates can cause these stable Gaussians to drift toward transient, single-view appearances introduced by distractors, thereby reducing consistency and weakening geometric stability.

To mitigate the influence of such single-view signals while preserving multi-view coherence, we adopt a sparse color update strategy. Specifically, $t$ denotes the optimization step within phase $k$, and color parameters $\Theta_{\text{color}}^{t,k}$ for phase $k$ are updated only once every $N$ iterations as:

\begin{equation}
\Theta_{\text{color}}^{t,k} \leftarrow
\begin{cases}
\Theta_{\text{color}}^{t,k} - \eta \nabla_{\Theta_{\text{color}}^{t,k}} \mathcal{L}^{(k)}_{\text{filter}}, & t \bmod N = 0, \\[3pt]
\Theta_{\text{color}}^{t,k}, & \text{otherwise},
\end{cases}
\label{sparsecolor}
\end{equation}

while geometry and opacity parameters continue to be optimized at every step.
Leveraging the reliable SfM initialization, this strategy reinforces stability throughout the filtering process, which aims to progressively suppress distractor signals while maintaining overall consistency rather than pursuing exact color reconstruction.


\subsection{Reconstruction Phase}
\noindent\textbf{Overview.} 
While the preceding Progressive Filtering Phases ($1 \le k < K$) suppress distractors and stabilize multi-view geometry, they intentionally sacrifice fine-grained appearance to prevent overfitting to transient content.  
The goal of the Reconstruction Phase ($k = K$) is therefore to refine this purified representation by reintroducing the standard 3DGS objective, enabling recovery of detailed textures, shading, and subtle color variations.

\subsubsection{Initialization \texorpdfstring{$\boldsymbol{(k=K)}$}{(k=K)}}
Initialization plays a critical role in 3DGS training, as the early optimization trajectory heavily influences the final reconstruction quality.
While conventional pipelines initialize parameters using raw SfM points, we instead initialize Gaussian parameters for the Reconstruction Phase $\Theta^{(K)}$ with the final output of the Progressive Filtering Phase, denoted as $\Theta^{(K-1)}$.
Since $\Theta^{(K-1)}$ has undergone $K-1$ phases of discrepancy-guided filtering and mask-based structural optimization, it exhibits improved multi-view consistent geometry without interference with transient contents.

These characteristics make $\Theta^{(K-1)}$ a far more reliable and stable initialization than SfM.
Thus, the Reconstruction Phase begins from this purified parameter set, providing an ideal scaffold for fine-grained appearance recovery.

\subsubsection{Training Objective \texorpdfstring{$\boldsymbol{(k=K)}$}{(k=K)}.}
In the Reconstruction Phase, we introduce the standard 3DGS loss to recover fine-grained appearance, while applying masking to prevent the distractor reactivation. To this end, we use the mask $M_v^{\star}$ from the final filtering phase $\Theta^{(K-1)}$, and define the reconstruction objective as:
\begin{equation}
\tilde{I}^{gt,\star}_v \;=\; M^{\star}_v \odot I^{gt}_v,\quad
\tilde{I}^{r,\star}_v \;=\; M^{\star}_v \odot I^{r}_v\!\big(\Theta^{(K)}\big).
\end{equation}
\begin{equation}
\mathcal{L}_{\text{rec}}
=
\sum_{v \in \mathcal{V}}
\mathcal{L}_{\text{3DGS}}(\tilde{I}^{r,\star}_v, \tilde{I}^{gt,\star}_v),
\end{equation}
which restores fine details while preserving distractor-free, structurally consistent representation established by filtering phases. Using $M_v^{\star}$ from $k{=}K{-}1$ and optimizing $\Theta^{(K)}$, the reconstruction phase prevents previously excised distractor regions from influencing optimization, ensuring stable refinement.

\section{Experiments}

\begin{table*}[t]
  \centering
  \renewcommand{\arraystretch}{1.2}
\caption{Quantitative comparison between PDF-GS~(Ours) and recent State-of-the-Art methods on the \textbf{NeRF On-the-go} dataset~\cite{nerf-on-the-go}.
For DeSplat~\cite{desplat}, we have reproduced the results indicated as *, where we report the best scores over five runs, with all metrics obtained using the officially released training and rendering implementation.
}
  \begin{adjustbox}{max width=\linewidth}
    \setlength{\tabcolsep}{2pt}
    \begin{tabular}{l *{7}{ccc}}
    \toprule
    \multirow{3}{*}{Method} & 
      \multicolumn{6}{c}{Low Occlusion} &
      \multicolumn{6}{c}{Medium Occlusion} &
      \multicolumn{6}{c}{High Occlusion} &
      \multicolumn{3}{c}{} \\
    \cmidrule(lr){2-7}\cmidrule(lr){8-13}\cmidrule(lr){14-19}
      & \multicolumn{3}{c}{Mountain}
      & \multicolumn{3}{c}{Fountain}
      & \multicolumn{3}{c}{Corner}
      & \multicolumn{3}{c}{Patio}
      & \multicolumn{3}{c}{Spot}
      & \multicolumn{3}{c}{Patio-High}
      & \multicolumn{3}{c}{\textbf{Mean}} \\
      \cmidrule(lr){2-4}\cmidrule(lr){5-7}\cmidrule(lr){8-10}\cmidrule(lr){11-13}\cmidrule(lr){14-16}\cmidrule(lr){17-19}\cmidrule(lr){20-22}
        & PSNR & SSIM & LPIPS
        & PSNR & SSIM & LPIPS
        & PSNR & SSIM & LPIPS
        & PSNR & SSIM & LPIPS
        & PSNR & SSIM & LPIPS
        & PSNR & SSIM & LPIPS
        & PSNR & SSIM & LPIPS \\
      \midrule
        3DGS~\cite{3dgs}
        & 19.22 & 0.69 & 0.23 & 20.08 & \thirdbest{0.69} & \bt{0.21}
        & 22.65 & 0.84 & 0.16 & 17.04 & 0.71 & 0.23
        & 18.54 & 0.72 & 0.33 & 17.04 & 0.66 & 0.31
        & 19.09 & 0.72 & 0.25 \\
        SpotLessSplats~\cite{spotless}
        & 20.67 & 0.67 & 0.28 & \thirdbest{20.63} & 0.65 & 0.27
        & 25.47 & 0.86 & 0.16 & \thirdbest{21.43} & 0.80 & 0.17
        & 23.64 & 0.82 & 0.21 & 21.17 & 0.75 & 0.24
        & 22.17 & 0.76 & 0.22 \\
        WildGaussians~\cite{wildgaussians}
        & \thirdbest{20.77} & 0.70 & 0.27 & 20.48 & 0.67 & 0.25
        & 25.21 & 0.87 & 0.14 & 21.17 & 0.80 & 0.17
        & 24.60 & 0.87 & 0.14 & \thirdbest{22.44} & 0.80 & 0.18
        & \thirdbest{22.45} & 0.78 & 0.19 \\
        DeSplat*~\cite{desplat}
        & 19.32 & \thirdbest{0.71} & \secondbest{0.20} & 20.45 & 0.68 & \thirdbest{0.21}
        & \thirdbest{26.30} & \thirdbest{0.88} & \thirdbest{0.11} & 18.81 & \thirdbest{0.81} & \best{0.14}
        & \secondbest{26.03} &\thirdbest{0.89} & \thirdbest{0.12} & 22.38 & \thirdbest{0.83} & \thirdbest{0.16}
        & 22.21 & \secondbest{0.80} & \thirdbest{0.16} \\
        RobustSplat~\cite{robustsplat}
        & \secondbest{21.15} & \secondbest{0.74} & \secondbest{0.20} & \secondbest{21.01} & \best{0.70} & \best{0.20}
        & \best{26.42} & \best{0.90} & \best{0.10} & \best{21.63} & \best{0.83} & \best{0.14}
        & \best{26.21} & \best{0.91} & \best{0.10} & \secondbest{22.87} & \best{0.84} & \best{0.15}
        & \secondbest{23.22} & \best{0.82} & \best{0.15} \\
        \midrule
        \textbf{PDF-GS~(Ours)}
        & \best{21.82} & \best{0.75} & \best{0.18} & \best{21.19} & \best{0.70} & \best{0.20}
        & \secondbest{26.41} & \best{0.90} & \best{0.10} & \secondbest{21.54} & \best{0.83} & \best{0.14}
        & \thirdbest{25.94} & \best{0.91} & \best{0.10} & \best{23.00} & \best{0.84} & \best{0.15}
        & \best{23.32} & \best{0.82} & \best{0.15} \\
      \bottomrule
    \end{tabular}
  \end{adjustbox}
  \label{tab:quantitative_onthego}
\end{table*}

\begin{table*}[t]
  \centering
  \renewcommand{\arraystretch}{1.2}
  \setlength{\tabcolsep}{2pt}
  
\caption{Quantitative comparison between PDF-GS~(Ours) and recent State-of-the-Art methods on the \textbf{RobustNeRF} dataset~\cite{robustnerf}. For DeSplat~\cite{desplat}, we have reproduced the results indicated as *, where we report the best scores over five runs, with all metrics obtained using the officially released training and rendering implementation.
}
  \begin{adjustbox}{max width=\linewidth}
    \begin{tabular}{l *{5}{ccc}}
      \toprule
      \multirow{2}{*}{Method} &
      \multicolumn{3}{c}{Android} &
      \multicolumn{3}{c}{Crab2} &
      \multicolumn{3}{c}{Statue} &
      \multicolumn{3}{c}{Yoda} &
      \multicolumn{3}{c}{\textbf{Mean}} \\
      \cmidrule(lr){2-4}\cmidrule(lr){5-7}\cmidrule(lr){8-10}\cmidrule(lr){11-13}\cmidrule(lr){14-16}
        & PSNR$\uparrow$ & SSIM$\uparrow$ & LPIPS$\downarrow$
        & PSNR$\uparrow$ & SSIM$\uparrow$ & LPIPS$\downarrow$
        & PSNR$\uparrow$ & SSIM$\uparrow$ & LPIPS$\downarrow$
        & PSNR$\uparrow$ & SSIM$\uparrow$ & LPIPS$\downarrow$
        & PSNR$\uparrow$ & SSIM$\uparrow$ & LPIPS$\downarrow$ \\
      \midrule
        3DGS~\cite{3dgs}                    & 23.32 & 0.79 & 0.16 & 31.76 & 0.93 & 0.17 & 20.83 & 0.83 & 0.15 & 28.92 & 0.91 & 0.19 & 26.21 & 0.86 & 0.17 \\
        WildGaussians~\cite{wildgaussians}  & \bs{24.67} & \bo{0.83} & \bt{0.15} & 30.52 & 0.91 & 0.21 & 22.54 & \bt{0.86} & 0.13 & 30.55 & 0.91 & 0.20 & 27.07 & \bt{0.88} & 0.17 \\
        SpotLessSplats~\cite{spotless}      & 24.20 & 0.81 & 0.16 & 33.90 & \bt{0.93} & 0.17 & 21.97 & 0.82 & 0.16 & \bt{34.24} & \bs{0.94} & 0.16 & 28.58 & \bt{0.88} & 0.16 \\
        DeSplat*~\cite{desplat}              & 24.20 & 0.82 & 0.16 & \bt{34.12} & \bt{0.93} & \bo{0.14} & \bs{22.93} & 0.85 & \bt{0.12} & 34.15 & \bs{0.94} & \bo{0.15} & \bt{28.85} & \bt{0.88} & \bt{0.15} \\
        RobustSplat~\cite{robustsplat}      & \bt{24.62} & \bo{0.83} & \bs{0.13} & \bs{34.88} & \bo{0.94} & \bs{0.15} & \bt{22.80} & \bo{0.87} & \bo{0.11} & \bs{35.14} & \bs{0.94} & \bo{0.15} & \bs{29.36} & \bo{0.90} & \bs{0.14} \\
        \midrule
        \textbf{PDF-GS~(Ours)}                & \bo{24.74} & \bo{0.83} & \bo{0.12} & \bo{35.35} & \bo{0.94} & \bs{0.15} & \bo{22.98} & \bo{0.87} & \bo{0.11} & \bo{35.54} & \bo{0.95} & \bo{0.15} & \bo{29.65} & \bo{0.90} & \bo{0.13} \\
      \bottomrule
    \end{tabular}
  \end{adjustbox}
  \label{tab:quantitative_robustnerf}
\end{table*}

\noindent\textbf{Implementation Details.}
We follow the 3DGS~\cite{3dgs} pipeline and adopt three progressive filtering phases followed by a reconstruction phase, resulting in a total of 40K optimization steps. Each filtering phase and the reconstruction phase are optimized for 10K steps. At the beginning of the reconstruction phase, all spherical harmonic coefficients above the zeroth order are re-initialized. During the first progressive filtering phase ($k$ = 1), the structure-oriented loss is rescaled to match the magnitude of the standard 3DGS loss used in the reconstruction stage. We dilate the discrepancy-based mask by 7 pixels prior to application.

\noindent\textbf{Experimental Setup.}
We evaluate our method on two challenging benchmarks: RobustNeRF~\cite{robustnerf} and NeRF On-the-go~\cite{nerf-on-the-go}, both of which contain substantial transient or view-inconsistent content. Following prior works~\cite{robustsplat, desplat}, all input images are downscaled by a factor of 8, with the exception of the \textit{patio} scene in NeRF On-the-go, for which we use a downscaling factor of 4.

\subsection{Quantitative Evaluation}
As shown in Tab.~\ref{tab:quantitative_onthego} and Tab.~\ref{tab:quantitative_robustnerf}, PDF-GS consistently outperforms prior methods in all three metrics (PSNR, SSIM, and LPIPS), leading to a new state-of-the-art performance. 
Specifically, PDF-GS surpasses prior approaches that rely on predicting masks with dedicated predictors~\cite{wildgaussians, spotless, robustsplat} as well as methods that decompose static content and distractors using per-view Gaussians~\cite{desplat}. This improvement largely stems from our multi-phase optimization, where the filtering phases effectively remove distractor-induced inconsistencies, and the subsequent reconstruction phase faithfully restores fine geometric and appearance details without reintroducing artifacts.

\subsection{Qualitative Evaluation}
In Fig.~\ref{fig:fig_qualitaive}, we show qualitative comparison of our method and previous methods. Our method effectively removes distractors while preserving fine-grained scene details and accurately rendering static regions, as highlighted by the red arrows. The progressive filtering phases suppress view-inconsistent content, and the reconstruction phase recovers detailed appearance, together yielding distractor-free, high-fidelity reconstructions.

\begin{figure}[h]
  \centering
  \includegraphics[width=0.8\columnwidth, trim=0 22 0 0, clip]{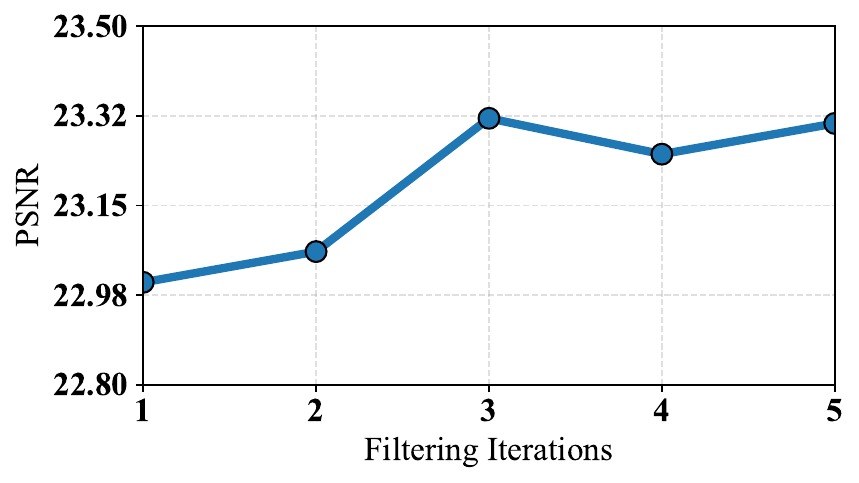}\par
\vspace{-6pt}
\hspace{0.11\columnwidth}\makebox[0.6\columnwidth]{\fontsize{7.5}{9}\selectfont Filtering Phases}
  \caption{\textbf{Analysis on the number of filtering phases.} 
    Quantitative results show a gradual increase in reconstruction scores as the number of filtering phases increases, with performance saturating around three phases (NeRF On-the-go dataset~\cite{nerf-on-the-go}).}
  \label{fig:ablation_num_phases}
  \vspace{-5pt}
\end{figure}
\begin{figure*}[t]
  \centering
  \includegraphics[width=\linewidth]{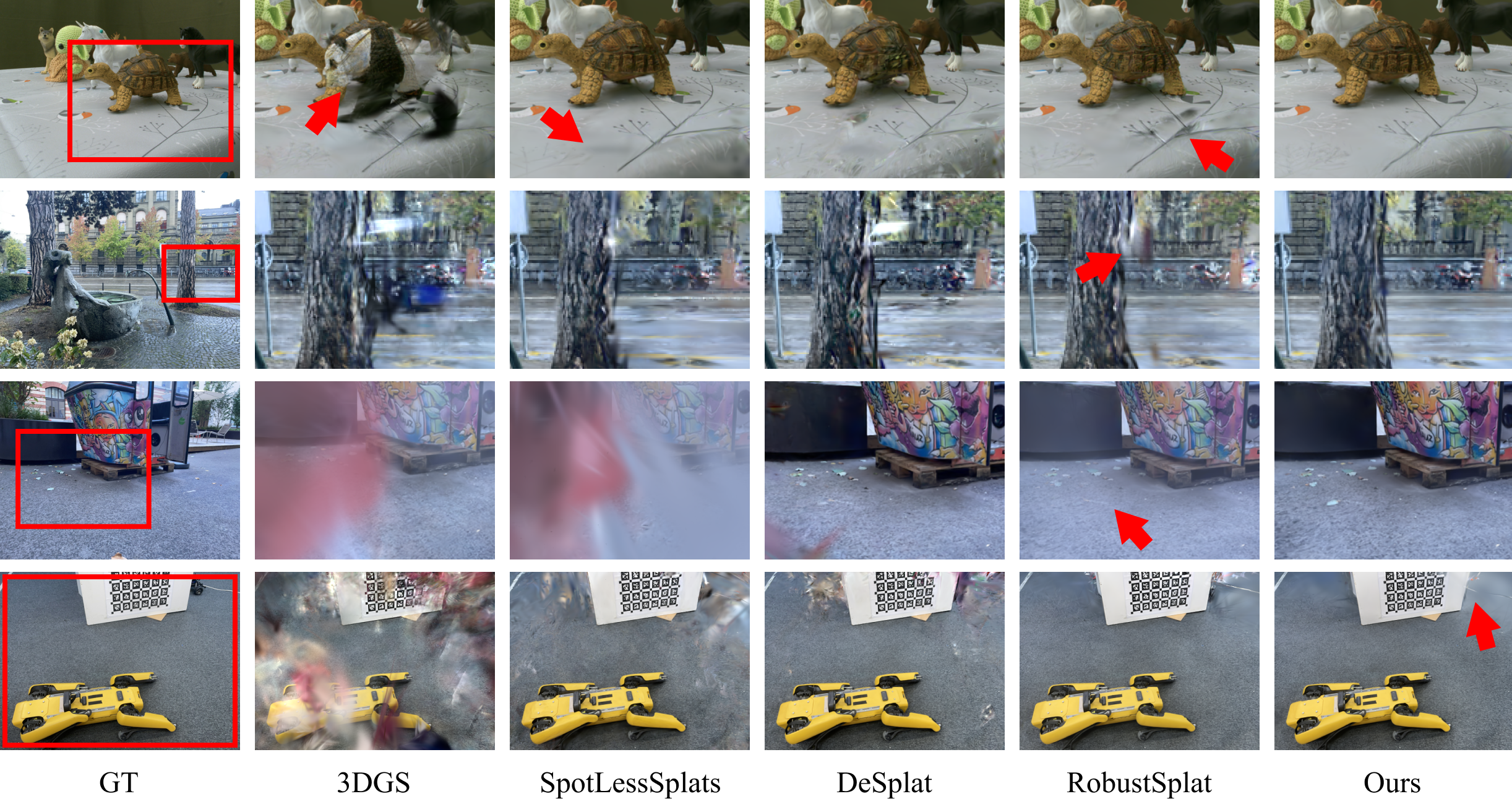}
  \caption{\textbf{Qualitative results of PDF-GS~(\textbf{Ours}) and baseline State-of-the-Art methods.} Our method generates noticeably fewer distractor-induced artifacts and more accurate reconstruction of static objects and backgrounds than previous methods.}
  \label{fig:fig_qualitaive}
\end{figure*}

\subsection{Ablation Study}

To better understand the contribution of each component in our framework, we conduct a series of ablation studies on the NeRF On-the-go dataset. Unless otherwise stated, we use three progressive filtering phases ($K$ = 4 including the final reconstruction phase) as the default configuration.

\subsubsection{Number of Progressive Filtering Phases.}
We evaluate the impact of the number of Progressive Filtering Phases on the final reconstruction quality (Fig.~\ref{fig:ablation_num_phases}).
As the number of filtering phases increases, performance consistently improves because each phase further removes residual distractors that remain from previous stages.

The most notable improvement is observed when increasing the number of phases from one to two, suggesting that a single filtering pass is insufficient for reliable distractor suppression. Beyond three phases, the performance gain saturates, as most distractor-prone regions have already been suppressed and subsequent passes mainly refine minor inconsistencies.

To balance quality and efficiency, we therefore adopt three filtering phases as the default setting for all experiments, achieving strong reconstruction quality while maintaining moderate training cost.

\subsubsection{Effect of Components on the Filtering Phase}

We then analyze the contribution of two key design choices in our filtering phase objective: (1) structure-only loss (Eq.~\ref{ssim} for $k{=}1$), and (2) sparse color update scheme (Eq.~\ref{sparsecolor}).

Results in Tab.~\ref{tab:ablation} show that removing either component degrades performance.
Without the structure-oriented loss, optimization easily overfits to local color inconsistencies and may fail to sufficiently suppress distractor artifacts. Without sparse color updates, Gaussian colors gradually drift toward single-view biases, weakening multi-view consistency.

Using both strategies jointly yields the highest reconstruction quality, validating that the combination of structure-guided supervision and controlled color updates effectively exposes and removes distractors while maintaining geometric stability.

\begin{table}[t]
\centering
\small
\renewcommand{\arraystretch}{1.2}
\setlength{\tabcolsep}{5pt}
\setlength{\aboverulesep}{2pt}
\setlength{\belowrulesep}{2pt}
\setlength{\abovetopsep}{2pt}
\setlength{\belowbottomsep}{2pt}

\caption{Ablation studies on the NeRF On-the-go dataset~\cite{nerf-on-the-go}. We ablate filtering phase components, re-initialization strategies, and threshold scheduling. $\dagger$ indicates our default setting.}
\begin{tabular}{l ccc}
  \toprule
  Setting & PSNR & SSIM & LPIPS \\
  \midrule
  \rowcolor{gray!15} \multicolumn{4}{l}{\textbf{\textit{Filtering Phase Components}}} \\
  Struct.\ Loss + Sparse Color$^\dagger$ & \bo{23.32} & \bo{0.82} & \bo{0.15} \\
  Struct.\ Loss only              & \bt{23.06} & \bo{0.82} & \bo{0.15} \\
  Sparse Color only               & \bs{23.20} & \bo{0.82} & \bo{0.15} \\
  Neither                          & \bn{23.02} & \bn{0.81} & \bo{0.15} \\
  \midrule
  \rowcolor{gray!15} \multicolumn{4}{l}{\textbf{\textit{Re-initialization Scheme}}} \\
  Between Filtering Phases$^\dagger$ & \bo{23.32} & \bo{0.82} & \bs{0.15} \\
  Always                             & \bs{23.08} & \bt{0.81} & \bt{0.16} \\
  Never                              & \bt{22.95} & \bo{0.82} & \bo{0.13} \\
  \midrule
  \rowcolor{gray!15} \multicolumn{4}{l}{\textbf{\textit{Threshold Schedule}}} \\
  Decreasing$^\dagger$  & \bo{23.32} & \bo{0.82} & \bo{0.15} \\
  Static              & \bs{23.06} & \bs{0.81} & \bo{0.15} \\
  \bottomrule
\end{tabular}
\label{tab:ablation}
\end{table}

\subsubsection{Gradually Decreasing Threshold.}

During the progressive filtering process, we progressively decrease the threshold $\tau_k$ used in Eq.~\ref{maskgen} to generate binary masks. That is, early phases conservatively flag only clear outliers with very large discrepancy as distractors, while subsequent phases apply stricter criteria, gradually excluding borderline regions that were previously tolerated.

As shown in Tab.~\ref{tab:ablation}, the proposed decreasing schedule outperforms a static threshold (23.32 vs.~23.06 dB), confirming that a progressive transition from conservative to strict masking yields a better balance between distractor suppression and detail preservation.

\begin{figure}[t]
  \centering
  \begin{subfigure}[t]{0.48\columnwidth}
    \centering
    \includegraphics[width=\linewidth, trim=10 50 220 100, clip]{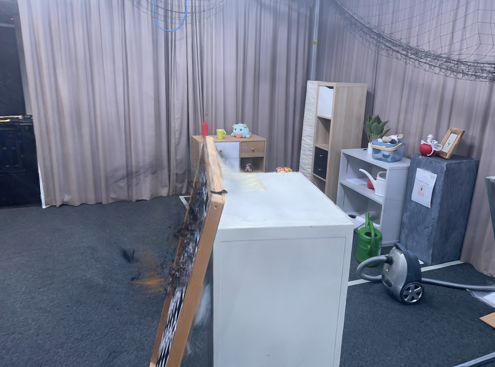}
    \caption{w/o Reinitialization}
  \end{subfigure}\hfill
  \begin{subfigure}[t]{0.48\columnwidth}
    \centering
    \includegraphics[width=\linewidth, trim=10 50 220 100, clip]{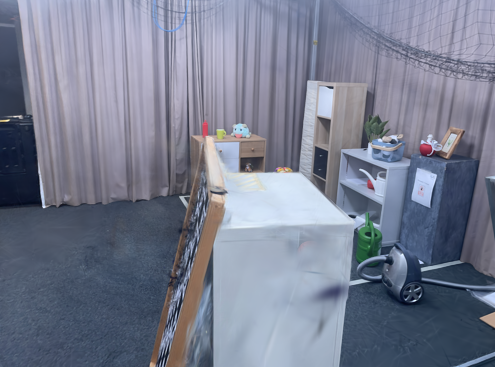}
    \caption{Ours}
  \end{subfigure}\hfill
    \caption{\textbf{Effect of re-initialization between filtering phases.} 
     \textbf{(a) Without re-initialization.} Accumulated errors propagate across filtering iterations and appear as persistent artifacts. \textbf{(b) With re-initialization (Ours).} Such error buildup is avoided, resulting in a cleaner and more stable reconstruction. (\textit{Corner} scene in NeRF On-the-go dataset~\cite{nerf-on-the-go}).
    }
  \label{fig:filtering}

\end{figure}

\subsubsection{Re-initialization Between Filtering Phases.}

We examine the effect of different Gaussian re-initialization strategies across filtering phases.
We compare three settings: (1) Re-initializing with SfM points at each filtering phase, and fine-tuning the final Gaussians from the last filtering phase during the reconstruction phase, (2) re-initializing before every phase, and (3) continuing optimization from previous parameters without re-initialization.

Tab.~\ref{tab:ablation} shows that our default setting achieves the best performance.
Always re-initializing leads to slower convergence and loss of accumulated geometry, while never re-initializing causes color drift and error accumulation.
Thus, periodic re-initialization effectively resets transient biases while preserving stable structures, leading to consistent and robust optimization across phases.

\begin{figure}[t]
    \centering

    \includegraphics[width=0.24\columnwidth, trim=0 0 0 100, clip]{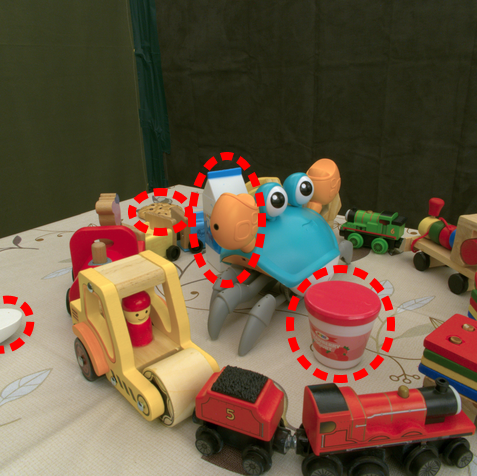}\hspace{3pt}%
    \includegraphics[width=0.24\columnwidth, trim=0 0 0 100, clip]{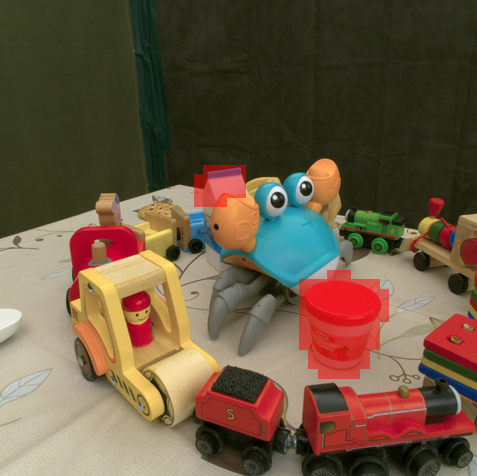}\hspace{3pt}%
    \includegraphics[width=0.24\columnwidth, trim=0 0 0 100, clip]{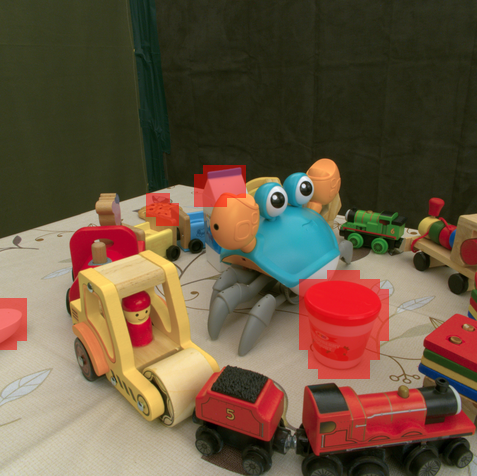}\hspace{3pt}%
    \includegraphics[width=0.24\columnwidth, trim=0 0 0 100, clip]{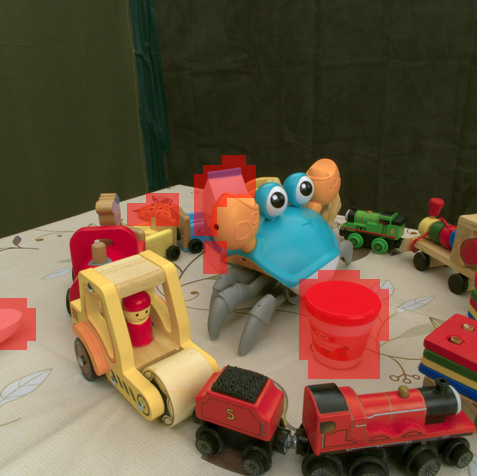}

    \vspace{4pt}

    \includegraphics[width=0.24\columnwidth]{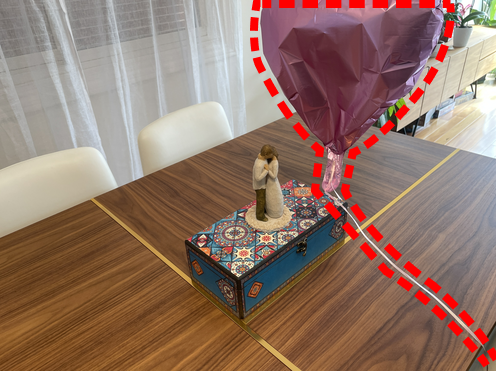}\hspace{3pt}%
    \includegraphics[width=0.24\columnwidth]{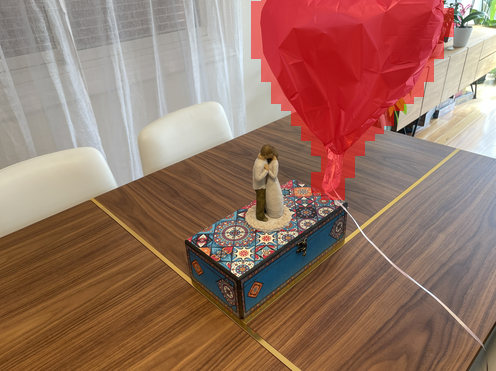}\hspace{3pt}%
    \includegraphics[width=0.24\columnwidth]{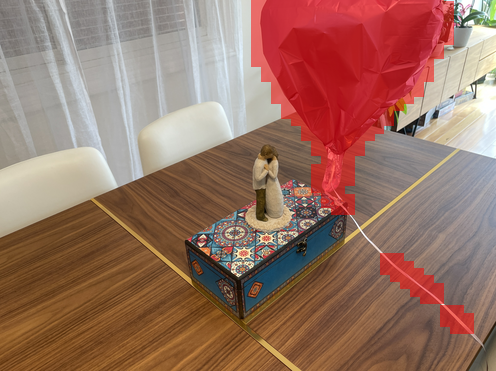}\hspace{3pt}%
    \includegraphics[width=0.24\columnwidth]{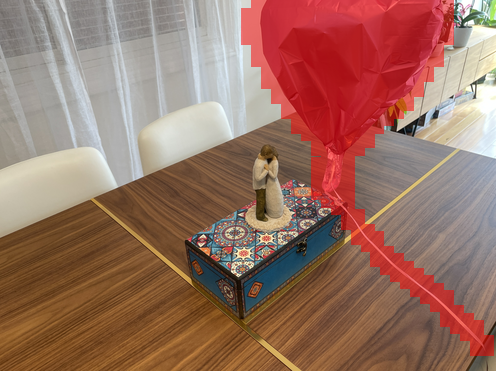}

    \vspace{4pt}

    \includegraphics[width=0.24\columnwidth]{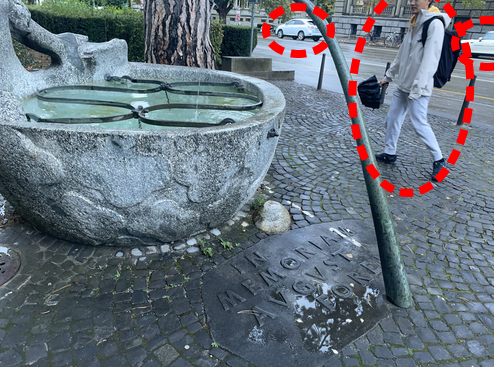}\hspace{3pt}%
    \includegraphics[width=0.24\columnwidth]{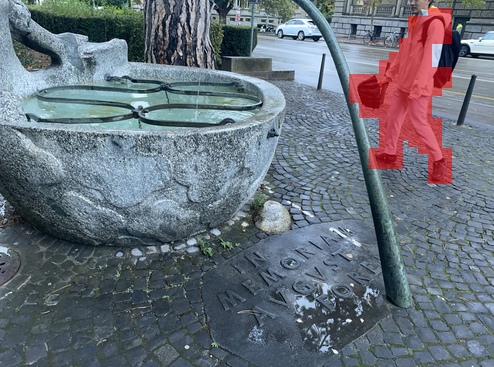}\hspace{3pt}%
    \includegraphics[width=0.24\columnwidth]{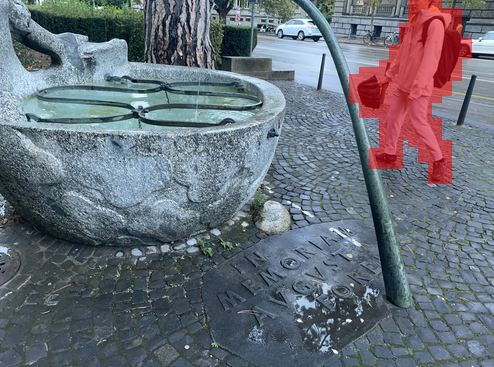}\hspace{3pt}%
    \includegraphics[width=0.24\columnwidth]{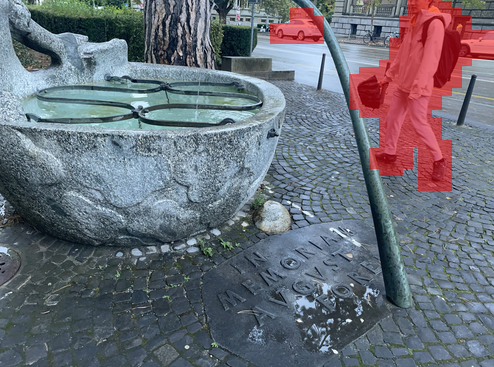}

    \vspace{4pt}

    \includegraphics[width=0.24\columnwidth]{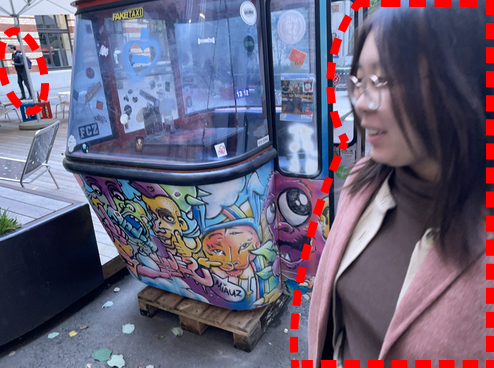}\hspace{3pt}%
    \includegraphics[width=0.24\columnwidth]{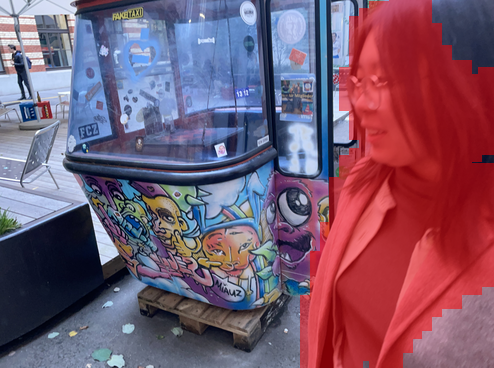}\hspace{3pt}%
    \includegraphics[width=0.24\columnwidth]{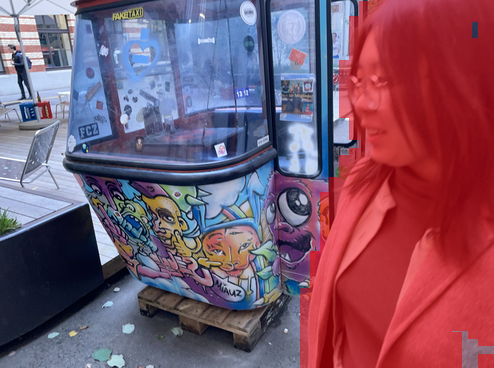}\hspace{3pt}%
    \includegraphics[width=0.24\columnwidth]{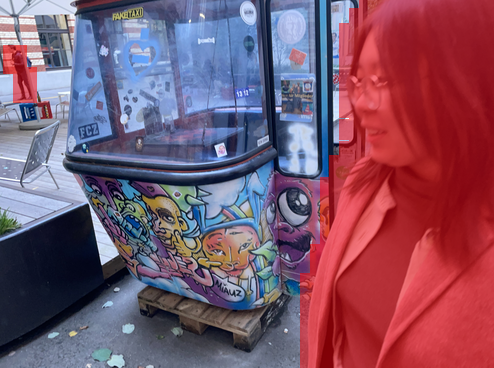}

    \vspace{2pt}

    \begin{minipage}{0.24\columnwidth}\centering Train View\end{minipage}\hspace{3pt}%
    \begin{minipage}{0.24\columnwidth}\centering Filtering Phase 1\end{minipage}\hspace{3pt}%
    \begin{minipage}{0.24\columnwidth}\centering Filtering Phase 2\end{minipage}\hspace{3pt}%
    \begin{minipage}{0.24\columnwidth}\centering Filtering Phase 3\end{minipage}
    
    \caption{
    \textbf{Progressive mask refinement across phases.}
    The dotted outlines in the images of the leftmost column denote the distractor region, and the red overlay indicates the masked pixels.
    As training progresses through phases, the masks become increasingly refined and better aligned with true distractors.
    For example, in the third row (\textit{fountain} scene from the NeRF On-the-go dataset~\cite{nerf-on-the-go}), background cars that were not masked in earlier phases are correctly identified as distractors in later phases.
    }
    \label{fig:mask_phase_visualization}
\end{figure}

\subsection{Visualization of Progressive Mask Evolution}

Fig.~\ref{fig:mask_phase_visualization} visualizes how per-view masks evolve across our progressive filtering phases.
In the first phase, distractors have not yet been removed, which leads to supervision that is noticeably noisier than in later stages. To avoid discarding valid content under this noisy setting, we adopt a conservative masking threshold that filters out only regions showing strong discrepancy signals.

As training moves through later phases, the supervision becomes progressively cleaner, allowing the masking criterion to be applied more strictly. The masks therefore become more effective over phases, enabling a larger portion of distractors to be identified while consistently preserving view-consistent regions.

This progressive refinement reflects the core principle of our framework: each phase benefits from the increasingly cleaner supervision produced by the preceding one, leading to continual improvements in mask quality.

\section{Conclusion}
We presented PDF-GS, a progressive filtering framework that enhances the robustness of 3D Gaussian Splatting in real-world, distractor-rich scenes. Instead of relying on explicit mask prediction or scene decomposition, PDF-GS leverages the inherent self-filtering property of 3DGS and amplifies it through iterative discrepancy-guided refinement. Across multiple filtering phases, the model progressively suppresses view-inconsistent distractors while preserving stable, view-consistent structures. A final reconstruction phase recovers fine-grained details from this purified representation, yielding high-fidelity and distractor-free results. Extensive experimental results show that PDF-GS consistently improves reconstruction quality under noisy training images with distractors.

\section*{Acknowledgements}
This work was supported in part by MSIT/IITP (No. RS-2022-II220680, RS-2020-II201821, RS-2019-II190421, RS-2024-00459618, RS-2024-00360227, RS-2024-00437633, RS-2024-00437102, RS-2025-25442569), MSIT/NRF (No. RS-2024-00357729), and KNPA/KIPoT (No. RS-2025-25393280).

{
    \small
    \bibliographystyle{ieeenat_fullname}
    \bibliography{main}
}

\appendix
\clearpage
\setcounter{page}{1}
\setcounter{table}{0}
\setcounter{figure}{0}
\maketitlesupplementary

\renewcommand{\thetable}{A\arabic{table}}
\renewcommand{\thefigure}{A\arabic{figure}}

\section{Additional Experiments}

\subsection{Phase-wise Reconstruction Quality.}

\begin{figure}[h]
  \centering
  \begin{subfigure}{0.49\columnwidth}
      \centering
      \includegraphics[width=\linewidth]{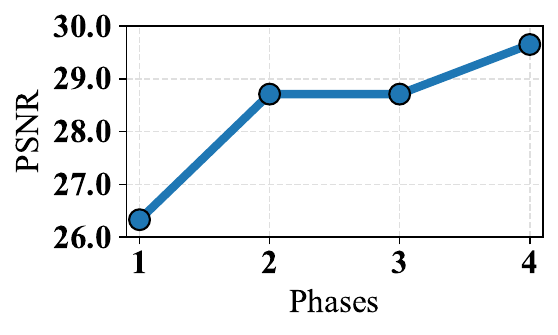}
      \caption{RobustNeRF dataset~\cite{robustnerf}}
      \label{fig:exp_phase_psnr_robustnerf}
  \end{subfigure}
  \hfill
  \begin{subfigure}{0.48\columnwidth}
      \centering
      \includegraphics[width=\linewidth]{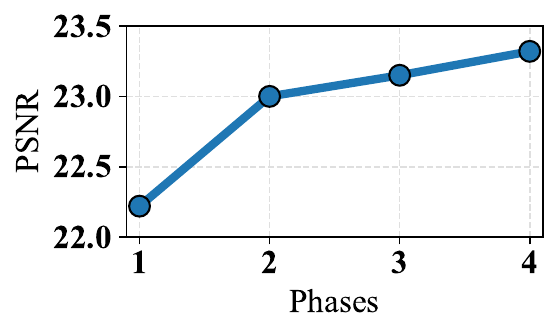}
      \caption{NeRF On-the-go dataset~\cite{nerf-on-the-go}}
      \label{fig:exp_phase_psnr_nerf_onthego}
  \end{subfigure}
  \caption{
    \textbf{Phase-wise evolution of reconstruction quality.}
    Both datasets show a gradual increase in reconstruction quality as the model advances through the phases of our progressive filtering framework, with improvements accumulating across successive phases.
  }
  \label{fig:exp_phase_psnr}
  \vspace{-5pt}
\end{figure}

In addition to evaluating the final performance for different numbers of filtering phases, we further analyze how reconstruction quality evolves during training by reporting the phase-wise PSNR averaged over all scenes in both the RobustNeRF~\cite{robustnerf} and NeRF On-the-go~\cite{nerf-on-the-go} datasets (Fig.~\ref{fig:exp_phase_psnr}).

Across both benchmarks, the PSNR consistently increases as the model progresses through successive phases. Each phase removes additional multi-view inconsistent signals, enabling cleaner supervision for the subsequent phases. Thus, the progressive design leads to cumulative quality improvements beyond single-pass optimization (i.e., training performed in one continuous optimization).

The first phase shows relatively low PSNR due to the use of a structure-oriented loss, which emphasizes local structures and reduces sensitivity to photometric cues affected by distractors. In addition, no mask is available at this stage and distractors have not yet been removed, resulting in considerably noisier supervision than in later phases. As training proceeds through the filtering phases, the masks and Gaussian representations are progressively refined, and by the final phase the model benefits from these improvements together with standard 3DGS objective, yielding the highest reconstruction quality.

\subsection{Progressive Filtering on Single-Pass Baselines}

\begin{table}[h]
\vspace{0pt}
\centering
\caption{
Evaluation of applying our progressive filtering procedure to a single-pass baseline (RobustSplat~\cite{robustsplat}) on the RobustNeRF dataset~\cite{robustnerf}. We report results using both DINOv2~\cite{dinov2} and DINOv3~\cite{dinov3} backbones, and $\ast$ indicates our reproduced results.}

\small
\renewcommand{\arraystretch}{1.1}
\setlength{\tabcolsep}{6pt}
\begin{tabular}{l ccc}
  \toprule
  Method & PSNR & SSIM & LPIPS \\
  \midrule
  RobustSplat$\ast$~\cite{robustsplat} (DINOv2) & 29.24 & 0.89 & 0.13 \\
  \ \ + Progressive Filtering (DINOv2) & 29.32 & 0.90 & 0.14 \\
  \midrule
  RobustSplat$\ast$~\cite{robustsplat} (DINOv3) & 29.14 & 0.89 & 0.13\\
  \ + Progressive Filtering (DINOv3) & 29.43 & 0.89 & 0.15 \\
  \bottomrule
\label{tab:sota_compatibility}
\end{tabular}
\end{table}

We examine how our progressive filtering strategy can improve methods that perform distractor removal and reconstruction in a single optimization pass, using RobustSplat~\cite{robustsplat} as a representative example. While such single-pass approaches aim to complete filtering and reconstruction in one continuous process, our framework conducts these steps progressively across multiple filtering phases followed by a reconstruction phase. This difference motivates evaluating whether progressive filtering can enhance a method originally designed for a single-pass pipeline.

Specifically, we augment RobustSplat~\cite{robustsplat} by running its training loop for three successive phases while keeping its original architecture unchanged.
Each phase is trained for 15k iterations, summing to 45k iterations in total, which is comparable to the 40k iterations used in our full framework. In each phase, we apply only the progressive filtering procedure, where phase-wise masks are computed from the discrepancy between the ground-truth images and the rendered training views of the preceding phase. Since RobustSplat~\cite{robustsplat} is designed to perform reconstruction within a single optimization pass, no additional reconstruction stage is introduced beyond these repeated optimization loops.

We observe this simple form of integration improves reconstruction quality (Tab.~\ref{tab:sota_compatibility}), indicating that our progressive filtering can strengthen the reconstruction process even when applied to methods originally designed for a single optimization pass.

\subsection{Integration of Learned Mask Predictors}

\begin{table}[h]
\vspace{0pt}
\centering
\caption{
Comparison of different masking strategies within our framework on the NeRF On-the-go dataset~\cite{nerf-on-the-go}.}

\small
\renewcommand{\arraystretch}{1.1}
\setlength{\tabcolsep}{6pt}
\begin{tabular}{lccc}
\toprule
Masking Method & PSNR & SSIM & LPIPS \\
\midrule
Discrepancy-based (Ours) & 23.32 & 0.82 & 0.15 \\
Learned predictor (RobustSplat) & 23.39 & 0.82 & 0.14 \\
\bottomrule
\end{tabular}
\label{tab:masking_module_comparison}
\end{table}

Beyond augmenting existing baselines with our progressive filtering strategy, we also investigate a complementary direction that incorporates mechanisms from previous work into the mask generation step, using RobustSplat~\cite{robustsplat} as a representative example.
Our framework uses a simple discrepancy-based masking scheme in conjunction with mechanisms that leverage and amplify the inherent tendency of 3DGS to suppress view-inconsistent signals during optimization. By feeding progressively cleaner supervision back into subsequent phases, this mechanism reinforces the natural filtering behavior of 3DGS. Despite its simplicity, this masking approach achieves strong performance.

At the same time, our framework is compatible with more sophisticated masking strategies, including those that incorporate learned predictors. We therefore integrate the masking strategy of RobustSplat~\cite{robustsplat}, which incorporates a learned MLP predictor, and use it as the masking component within our multi-phase pipeline.
As shown in Tab.~\ref{tab:masking_module_comparison}, employing RobustSplat's strategy within our framework further improves reconstruction quality, illustrating the complementary nature of our method and its compatibility with masking mechanisms developed in prior works.

\subsection{Training Speed Comparison}

\begin{table}[h]
\vspace{0pt}
\centering
\caption{
Wall clock training time comparison on the RobustNeRF~\cite{robustnerf} dataset.
}

\small
\renewcommand{\arraystretch}{1.1}
\setlength{\tabcolsep}{4pt}
\setlength{\aboverulesep}{2pt}
\setlength{\belowrulesep}{2pt}
\setlength{\abovetopsep}{2pt}
\setlength{\belowbottomsep}{2pt}

\begin{tabular}{l cccc}
  \toprule
  Method & Android & Crab2 & Statue & Yoda \\
  \midrule
  RobustSplat~\cite{robustsplat} & 21.6 min & 24.5 min & 28.7 min & 24.2 min \\
  Ours & 25.2 min & 23.5 min & 28.1 min & 23.9 min \\
  \bottomrule
\end{tabular}

\label{tab:train_speed}
\end{table}

We compare the wall clock training time of our method with RobustSplat~\cite{robustsplat} on the RobustNeRF~\cite{robustnerf} dataset, measuring end-to-end training time for each scene under the same hardware setting (Tab.~\ref{tab:train_speed}). 
Although our method uses 40k optimization iterations, which is more than the 30k iterations in RobustSplat, the overall training time remains comparable. This is largely because RobustSplat~\cite{robustsplat} performs feature extraction at every iteration, whereas our method computes features only between phases. This per-phase design decouples feature extraction from the inner optimization loop, allowing it to be performed significantly less frequently and making it feasible to use heavier features or more advanced techniques. Exploring such extensions is an interesting direction for future work.

\subsection{Effect on Different Masking Metrics.}

\begin{table}[t]
\vspace{0pt}
\centering

\small
\renewcommand{\arraystretch}{1.1}
\setlength{\tabcolsep}{4pt}
\setlength{\aboverulesep}{2pt}
\setlength{\belowrulesep}{2pt}
\setlength{\abovetopsep}{2pt}
\setlength{\belowbottomsep}{2pt}
\caption{Quantitative results with different masking metrics. $\dagger$ indicates our default setting. Our method with DINOv3 yields the best performance. Even when using low-level metrics such as PSNR and SSIM, our method outperforms vanilla 3DGS. }
\begin{tabular}{l ccc}
  \toprule
  Method & PSNR & SSIM & LPIPS \\
  \midrule
  DINOv3$^\dagger$ & \bo{29.65} & \bo{0.90} & \bo{0.13} \\
  DINOv2           & \bs{29.42} & \bs{0.89} & \bt{0.14} \\
  SSIM             & \bt{28.88} & \bs{0.89} & \bo{0.13} \\
  PSNR             & \bn{28.71} & \bs{0.89} & \bt{0.14} \\
  \bottomrule
\end{tabular}
\label{tab:ablation_noisetype}
\vspace{-8pt}
\end{table}

We next study the effect of different feature transformations $F(\cdot)$ in Eq.~\ref{Eq: discrepancy metric} of the main paper for computing the discrepancy map between rendered and ground-truth images. This choice directly affects how distractor regions are localized and masked out.

As shown in Tab.~\ref{tab:ablation_noisetype},  DINOv3 features yield the best overall performance (PSNR = 29.65 dB), outperforming earlier versions such as DINOv2 (29.42 dB) and simple low-level metrics like SSIM or PSNR. However, note that even when low-level metrics as PSNR and SSIM are employed (i.e., without any pretrained model), our method consistently surpasses vanilla 3DGS across all evaluation metrics.
This confirms that the proposed progressive filtering strategy itself is intrinsically effective, while stronger feature representations such as DINOv3 further amplify its robustness and accuracy.


\end{document}